\documentclass[unnumsec,webpdf,contemporary,large]{oup-authoring-template}%
\usepackage{hyperref}
\hypersetup{
    colorlinks=true,       
    linkcolor=blue,        
    citecolor=magenta,       
    filecolor=green,     
    urlcolor=cyan          
}

\theoremstyle{thmstyleone}%
\theoremstyle{thmstyletwo}%
\theoremstyle{thmstylethree}%

\begin{document}

\journaltitle{Journal Title Here}
\DOI{DOI HERE}
\copyrightyear{2022}
\pubyear{2019}
\access{Advance Access Publication Date: Day Month Year}
\appnotes{Paper}

\firstpage{1}


\title[w1ot]{Fast and scalable Wasserstein-1 neural optimal transport solver for single-cell perturbation prediction}

\author[1,2]{Yanshuo Chen\ORCID{0000-0001-6675-2449}}
\author[1]{Zhengmian Hu\ORCID{0000-0003-0316-146X}}
\author[3,4]{Wei Chen}
\author[1,2]{Heng Huang}

\authormark{Chen et al.}

\address[1]{\orgdiv{Department of Computer Science}, \orgname{University of Maryland}, \orgaddress{\street{College Park}, \postcode{20742}, \state{MD}, \country{USA}}}

\address[2]{\orgdiv{Center of Bioinformatics and Computational Biology}, \orgaddress{\postcode{20740}, \state{MD}, \country{USA}}}
\address[3]{\orgdiv{Department of Biostatistics}, \orgname{University of Pittsburgh}, \orgaddress{\postcode{15261}, \state{PA}, \country{USA}}}
\address[4]{\orgdiv{Department of Pediatrics}, \orgname{UPMC Children's Hospital of Pittsburgh}, \orgaddress{\postcode{15224}, \state{PA}, \country{USA}}}

\corresp[$\ast$]{Corresponding author. \href{email:email-id.com}{heng@umd.edu}}

\received{Date}{0}{Year}
\revised{Date}{0}{Year}
\accepted{Date}{0}{Year}



\abstract{\textbf{Motivation:}
Predicting single-cell perturbation responses requires mapping between two unpaired single-cell data distributions. Optimal transport (OT) theory provides a principled framework for constructing such mappings by minimizing transport cost. Recently, Wasserstein-2 ($W_2$) neural optimal transport solvers (\textit{e.g.}, CellOT) have been employed for this prediction task. However, $W_2$ OT relies on the general Kantorovich dual formulation, which involves optimizing over two conjugate functions, leading to a complex min-max optimization problem that converges slowly. \\
\textbf{Results:}
To address these challenges, we propose a novel solver based on the Wasserstein-1 ($W_1$) dual formulation. Unlike $W_2$, the $W_1$ dual simplifies the optimization to a maximization problem over a single 1-Lipschitz function, thus eliminating the need for time-consuming min-max optimization. While solving the $W_1$ dual only reveals the transport direction and does not directly provide a unique optimal transport map, we incorporate an additional step using adversarial training to determine an appropriate transport step size, effectively recovering the transport map. Our experiments demonstrate that the proposed $W_1$ neural optimal transport solver can mimic the $W_2$ OT solvers in finding a unique and ``monotonic" map on 2D datasets. Moreover, the $W_1$ OT solver achieves performance on par with or surpasses $W_2$ OT solvers on real single-cell perturbation datasets. Furthermore, we show that $W_1$ OT solver achieves $25 \sim 45\times$ speedup, scales better on high dimensional transportation task, and can be directly applied on single-cell RNA-seq dataset with highly variable genes. \\
\textbf{Availability and Implementation:}
Our implementation and experiments are open-sourced at \url{https://github.com/poseidonchan/w1ot}. \\
\textbf{Contact:}
Yanshuo Chen (\url{cys@umd.edu}) and Heng Huang (\url{heng@umd.edu}). \\
\textbf{Supplementary Information:} 
Supplementary data are available online at the journal's website.}

\keywords{Perturbation, Single-cell data, Wasserstein-1 optimal transport}

\maketitle

\section{Introduction}
Single-cell perturbation experiments have emerged as a powerful tool for modeling cellular responses to environmental changes, such as drug treatments and other stimuli. These experiments are crucial for characterizing how cells react to perturbations, which has significant implications for fields such as cancer research \citep{cancer} and drug development \citep{drug}. Common approaches for single-cell perturbation experiments include single-cell RNA sequencing (scRNA-seq) \citep{scperturb} and Iterative Indirect Immunofluorescence Imaging (4i) \citep{4i}. scRNA-seq allows for the measurement of gene expression changes in response to genetic \citep{norman} or drug \citep{sciplex} perturbations, while 4i provides spatially resolved information on protein abundance and localization.

Despite the valuable insights these techniques provide, they also introduce challenges in data analysis. In these experiments, cells are typically fixed or destroyed during measurement, making it impossible to observe the same cell before and after perturbation. Consequently, modeling and predicting single-cell perturbation responses becomes a challenging task due to the inherent unpaired nature of the data. Researchers must work with separate distributions of control and perturbed cells, presenting a complex problem of matching cells between conditions while accounting for cellular heterogeneity.

Several methods have been proposed to address the challenge of predicting single-cell perturbation responses. The baseline method, scGen \citep{scgen}, calculates the shift between the means of embedded target and source cells and applies it uniformly to transport the data. Building on this idea of applying a shift in the latent space, other methods aim to learn disentangled representations to model perturbation responses. For instance, chemCPA \citep{CPA,chemCPA} employs an encoder-decoder structure to learn disentangled representations of cells and drugs, allowing for additive operations in the latent space to predict perturbation effects. Similarly, biolord \citep{biolord} uses latent optimization techniques to better model different attributes. Although these methods can capture nonlinear responses, they transform the perturbation prediction problem into the potentially more challenging task of learning invariant and disentangled representations.

Recently, computational methods based on optimal transport (OT) theory — such as CellOT \citep{cellot} and CINEMA-OT \citep{cinemaot} — have offered a principled framework for mapping between unpaired distributions. However, CINEMA-OT is designed to model existing perturbation data with a discrete OT formulation and thus cannot be used for prediction tasks. On the other hand, CellOT is built on the continuous Wasserstein-2 ($W_2$) OT formulation \citep{makkuva2020optimal} and can be used for generative prediction. Nevertheless, the $W_2$ formulation requires optimizing a complex min-max problem and optimizing input convex neural networks (ICNNs) \citep{icnn}, which leads to slow convergence and poor performance on high-dimensional data, respectively \citep{w2benchmark}.

To address the computational and scalability limitations of existing $W_2$ OT solvers, we propose a novel solver based on the Wasserstein-1 ($W_1$) formulation. This approach offers two key advantages. First, the $W_1$ dual simplifies the min-max optimization problem over two convex functions into a maximization problem over a single 1-Lipschitz function \citep{villani2009optimal,computationalOT}, thereby avoiding the complexities inherent in min-max optimization. Second, by eliminating the need to parameterize convex functions using ICNNs, our approach achieves better scalability, as ICNNs can be difficult to optimize in high-dimensional settings \citep{w2benchmark}. 

In this paper, we develop the first accurate $W_1$ OT solver leveraging recent advancements in 1-Lipschitz neural networks \citep{groupsort,groupsort_nn} and generative adversarial networks (GANs) \citep{gan}. We demonstrate that our solver is capable of 1) finding the ``monotonic" transport map on 2 dimensional (2D) datasets; 2) achieving similar or better performance on real single-cell perturbation datasets; 3) being $25 \sim 45\times$ faster and more scalable on high dimensional datasets.

\section{Materials and methods}\label{methods}
\subsection{Method overview}

\begin{figure*}[!t]
\centering
\includegraphics[width=\textwidth]{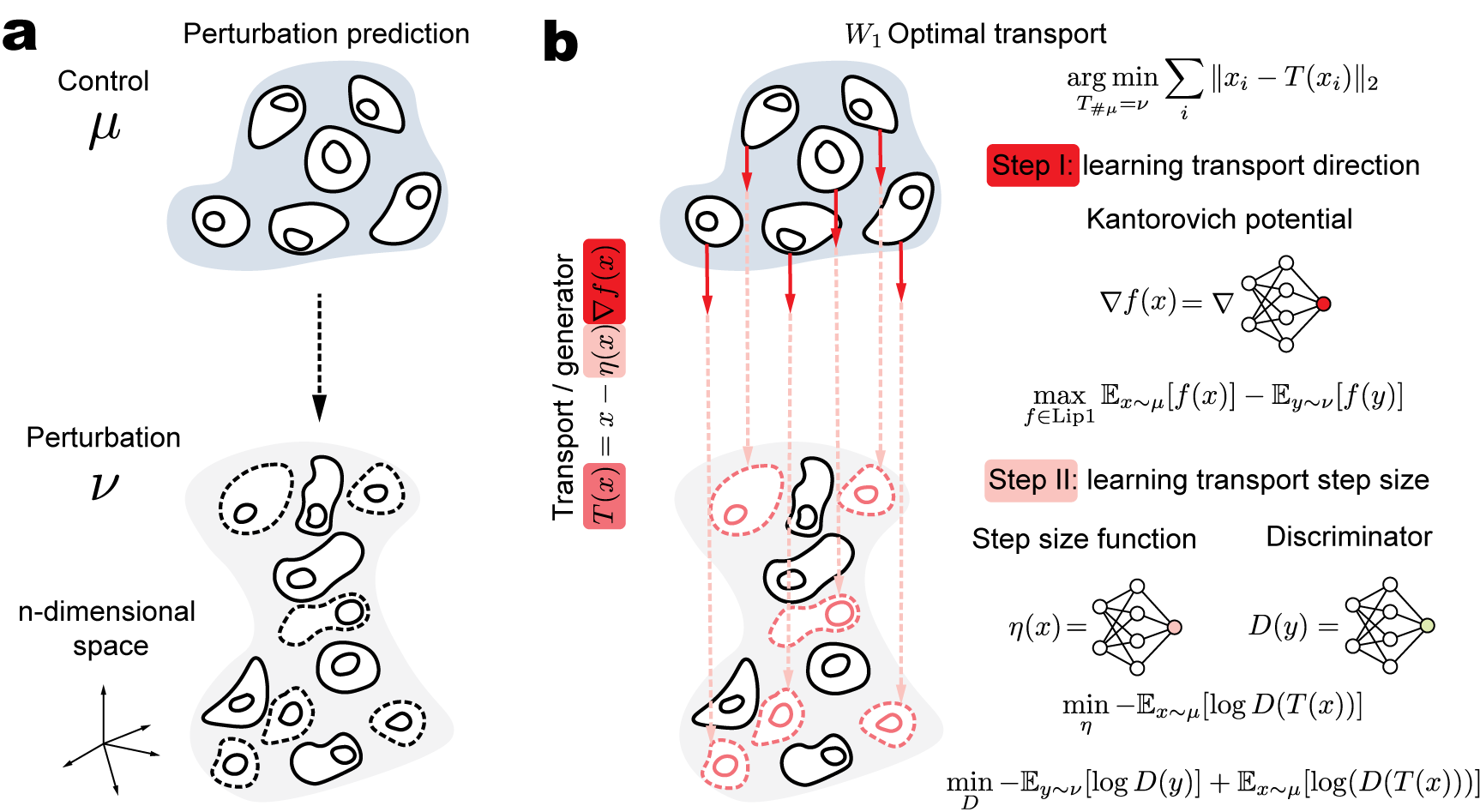}
\caption{\textbf{Overview of $W_1$OT} \textbf{a.} The single-cell perturbation prediction task needs a mapping to map the control group ($\mu$) cells to the perturbation group ($\nu$) cells. \textbf{b.} Our proposed $W_1$ OT solver. The first step is to learn the transport direction by maximizing the Kantorovich dual problem. The second step is to learn the appropriate transport step size via adversarial training to finally construct the transport map.Figure conceptually inspired by the illustration in the CellOT paper \cite{cellot}.} \label{fig1}
\end{figure*}

The single-cell perturbation task requires finding an unknown underlying map to transform the distribution of control group cells ($\mu$) to the distribution of perturbation group cells ($\nu$) in a high-dimensional space (Fig. \ref{fig1}a). Most current single-cell measurements destroy the cells, resulting in unpaired single-cell perturbation datasets. Optimal transport is a principled approach to map these unpaired data distributions. It assumes cells change following a ``minimal effort" principle, which has proven effective in many single-cell modeling problems \citep{ot_review}. Previous generative OT map models the transport effort as quadratic cost and has limitations on optimization complexity and scalability \citep{makkuva2020optimal,cellot,w2benchmark}. Since these limitations arise from the use of the $W_2$ Kantorovich dual, we instead develop an OT solver based on the $W_1$ dual (also known as the Kantorovich-Rubinstein dual) to avoid the min-max problem \citep{villani2009optimal}. However, building a $W_1$ OT solver presents theoretical challenges, as solving the dual problem does not directly recover the primal problem of finding the OT map; instead, it only provides insight into the transportation direction \citep{trudinger2001monge,caffarelli2002constructing}.  
Based on these analyses, our proposed $W_1$ OT solver learning procedure is divided into two steps: it first learns the transport direction $\nabla f$ by maximizing the $W_1$ dual, and then it learns the transport step size $\eta$ based on the transport direction guidance (Fig. \ref{fig1}b). We parameterize the 1-Lipschitz Kantorovich potential $f$ as a GroupSort neural network \citep{groupsort} and recover the transport direction with its gradient $\nabla f$. After fixing the transport direction, we parameterize the step size function $\eta$ as a non-negative deep neural network (DNN) and employ an additional DNN discriminator to adversarially train it to learn the appropriate step size. More background and details are included below.

\subsection{Methodology background}

\subsubsection{Wasserstein-1 optimal transport}
In optimal transport, we aim to find the optimal transport plan to transport one distribution to another in a sense of minimum total cost. If the cost is defined as the conventional distance metric which satisfies the triangular inequality, it is called Wasserstein-1 optimal transport. If the cost is defined as the distance metric to the power of $p$, then the problem is called as the Wasserstein-$p$ optimal transport. Compared to the Wasserstein-$p$ metric, the Wasserstein-1 optimal transport will be more robust to the outlier and the noise, and its elegant dual form will be numerically easy to solve \citep{computationalOT}. Hence, we focus on discussing the Wasserstein-1 optimal transport and its practical solver below.

\paragraph{Monge's optimal transport problem and Kantorovich duality}
Let $\mu$ and $\nu$ be two probability measures defined on measurable spaces $\mathcal{X} $ and $\mathcal{Y}  $, respectively. The cost function $c: \mathcal{X} \times \mathcal{Y} \rightarrow \mathbb{R}^+ $ represents the cost of moving a unit of mass from $x \in \mathcal{X}$ to $y \in \mathcal{Y}$. Monge's problem seeks to find a map $T$ that pushes $\mu$ forward to $\nu$ (\textit{i.e.}, $T_{\#} \mu = \nu$) while minimizing the total transport cost:

\begin{align}
\inf_{T_{\#} \mu = \nu} \int_\mathcal{X} c(x, T(x))\ d\mu(x)
\end{align}
However, Monge's problem may not always have a solution, particularly when $\mu$ has atomic mass (\textit{i.e.}, Dirac delta distribution) and $\nu$ does not. To address this, Kantorovich relaxed the problem by considering joint probability measures $\pi$ on $\mathcal{X} \times \mathcal{Y}$ with marginals $\mu$ and $\nu$, leading to the Kantorovich formulation:

\begin{align}
\inf_{\pi \in \Pi(\mu, \nu)} \int_{\mathcal{X} \times \mathcal{Y}} c(x, y)\ d\pi(x, y)
\end{align}
The solution of this problem is described as a transport plan instead of a transport map. The dual of this problem, known as the Kantorovich dual problem, is:

\begin{align} \label{eq:3}
\sup_{(\phi, \psi) \in \mathrm{\Phi}_c} \int_\mathcal{Y} \phi(y)\ d\nu(y) - \int_\mathcal{X} \psi(x)\ d\mu(x)
\end{align}
where $\mathrm{\Phi}_c = {(\phi, \psi) : \phi(y) - \psi(x) \leq c(x, y), \forall x \in \mathcal{X}, y \in \mathcal{Y}}$. To achieve the supremum (\textit{i.e.}, optimal cost or the Wasserstein distance), the functions $\psi$ and $\phi$ are related by the $c$-transform: $\phi(y) = \psi^c(y) = \inf_x (c(x, y) + \psi(x))$. When the cost function is the distance metric, the Wasserstein-1 distance $W_1(\mu, \nu)$ can be expressed using the Kantorovich-Rubinstein duality formula \citep{villani2009optimal}:

\begin{align} \label{eq:4}
W_1(\mu, \nu) = \sup_{f \in \text{Lip}1} \left( \int_\mathcal{X} f(x)\ d\mu(x) - \int_\mathcal{Y} f(y)\ d\nu(y) \right)
\end{align}
where $f$ is constrained to be a 1-Lipschitz continuous function and is referred as Kantorovich potential function in literature. Notably, the dual formulation of the Wasserstein-1 distance significantly simplifies the optimization procedure since we only need to optimize over a single function. In contrast, solving for the Wasserstein-$p$ ($p > 1$) is generally more challenging and often involves a min-max optimization problem \citep{makkuva2020optimal,cellot}.

\paragraph{Non-unique optimal transport map in 1-D segements} Unlike the Wasserstein-$p$ ($p>1$) scenario, where the optimal transport map can be uniquely determined by the gradient of the Kantorovich potential (i.e., Brenier's theorem \citep{brenier}) due to the strictly convex cost function, Wasserstein-1 Kantorovich potential only reveals the transport direction and its optimal transport has non-unique solutions due to the undetermined step size in the general setting \citep{gangbo_geometry_1996,caffarelli2002constructing,trudinger2001monge}:
\begin{align} \label{eq:5}
    T(x) = x - \eta(x) \frac{\nabla f(x)}{\| \nabla f(x) \|}
\end{align}
where $\| \cdot \|$ represents the Euclidean norm and $\eta(x)$ is the undetermined step size function. Of note, the norm of the gradient of the Kantorovich potential would be 1 when the optimal cost is achieved \citep{wgan}.

For example, imagine we have $n$ identical books lined up on a bookshelf, occupying positions 1 through $n$. If we want to optimally transport these books so they occupy positions 2 through $n+1$, there are multiple valid ways to do this. One solution could be to move the first book directly to position $n+1$, while another could be to shift each book one position to the right. Both solutions have the same total cost of $n$. We can see that in the Wasserstein-1 optimal transport setting, we only know the optimal transport direction but lack information on the transported distance for each point, which gives us multiple options to construct the final optimal transport map. Moreover, previous studies have shown that the non-uniqueness 1-D phenomenon is the only source of non-uniqueness in Monge’s problem \citep{feldman2002uniqueness}.

\paragraph{The unique and monotonic optimal transport map}
Among the various possible optimal transport maps, researchers aim to construct one that mimics the Wasserstein-$p$ optimal transport map and preserves the desired ``monotonicity" property. Returning to the previously introduced 1-D example, we might prefer the second optimal transport map because it ``monotonically" shifts all books to the next position, thereby preserving the local structure. Similarly, when mapping one single-cell data distribution to another, the desired transport is expected to maintain the local structure (\textit{e.g.}, cell type geometric structure) to preserve the biological meaning. Theoretically, previous geometric analysis has shown that this map exists, is unique, and should satisfy the following criterion \citep{caffarelli2002constructing,feldman2002uniqueness}:
\begin{align} \label{eq:6}
\frac{x_1 - x_2}{\|x_1-x_2\|} + \frac{T(x_1) - T(x_2)}{\|T(x_1)-T(x_2)\|} \neq 0
\end{align}
where $x_1 \neq x_2 \in \mathcal{X}$ and $T(x_1) \neq T(x_2)$. Furthermore, it has been shown that the inversion of (\ref{eq:6}) occurs only when $x_1, x_2, T(x_1), T(x_2)$ are collinear (i.e., $x_1$ and $x_2$ share the same transport direction, ${\nabla f(x_1)} = {\nabla f(x_2)}$) \citep{feldman2002uniqueness}.
This condition indicates that when they are collinear, the direction of the vector $x_1 - x_2$ is the same as the direction of the vector $T(x_1) - T(x_2)$. In other words, the transported points should maintain their relative order as in the source distribution.

\subsubsection{Wasserstein-2 optimal transport}
When the cost function is the square of a distance metric, the optimal transport problem is called Wasserstein-2 optimal transport problem. In this scenario, the Kantorovich duality (\ref{eq:3}) still holds and it can be transformed into a min-max optimization problem over convex functions \citep{makkuva2020optimal}:
\begin{align}\label{eq:7}
    W_2^2(\mu, \nu) = \sup_{\phi \in \text{CVX}(\mathcal{Y})} \inf_{\psi \in \text{CVX}(\mathcal{X})} - \int_\mathcal{Y} \phi(y)\ d\nu(y) \notag \\ -  \int_\mathcal{X} \langle x,\nabla \psi(x)\rangle - \phi(\nabla \psi(x))  \ d\mu(x) + C_{\mathcal{\mu}, \mathcal{\nu}} 
\end{align}
where $\text{CVX}(\cdot)$ represents the convex function defined on the corresponding space, $\langle \cdot, \cdot \rangle$ denotes the inner product operation, and $C_{\mu, \nu} = 1/2\ \mathbb{E}_\mathcal{Y} \| y \|^2 + 1/2 \ \mathbb{E}_\mathcal{X} \| x \|^2 $ is a constant. Under this formulation, the complicated $c$-transform relationship constraint between $\phi$ and $\psi$ is replaced by the constraint of convex function and the primal solution could be uniquely determined via the Brenier’s theorem \citep{brenier}:
\begin{align}
    T(x) = \nabla \psi(x)
\end{align}
Based on the elegant and unique solution guaranteed by the Brenier's theorem and the recently developed input convex neural network (ICNN) \citep{icnn}, OT-ICNN \citep{makkuva2020optimal} is proposed as a practical  Wasserstein-2 optimal transport solver with solid theoretical foundation and CellOT \citep{cellot} is developed based on it.

\subsubsection{1-Lipschitz neural network}
The 1-Lipschitz condition appears in the Kantorovich-Rubinstein dual can be defined as:
\begin{align}
    \| f(x) - f(y) \| \leq \| x - y \|
\end{align}
To incorporate this condition into a traditional deep neural network which is composed of $L$ different layers with activation function $\sigma$, it suffices to ensure each layer's transformation is 1-Lipschitz. For the $l$-th hidden layer $ \mathbf{h}_l$ with weight matrix $\mathbf{W}_l$ and bias $\mathbf{b}_{l}$:
\begin{align}
    \mathbf{h}_l = \sigma(\mathbf{W}_l \mathbf{h}_{l-1} + \mathbf{b}_{l})
\end{align}
We need to ensure that both the activation function $\sigma$ and the linear transformation are 1-Lipschitz.
\paragraph{1-Lipschitz linear transformation}
To ensure that the linear transformation is 1-Lipschitz, it is equivalent to ensure the weight matrix norm is less or equal than 1:
\begin{align}
\| \mathbf{W}_l \|_2 \leq 1
\end{align}
This condition ensures that the linear transformation does not amplify the input vectors, thereby maintaining the 1-Lipschitz property:
\begin{align}
\| \mathbf{W}_l \mathbf{h}_{l-1} \|_2 &\leq \| \mathbf{W}_l \|_2 \| \mathbf{h}_{l-1} \|_2 \leq \| \mathbf{h}_{l-1} \|_2
\end{align}

When backpropagating through norm-constrained networks, the gradient norms often decrease at each layer as the norm of weight matrix is smaller than 1. This vanishing gradient can be problematic when modeling functions that require preserved input-output gradient norms. For instance, in Wasserstein-1 distance estimation (\ref{eq:4}), the optimal dual function has a gradient norm of 1 almost everywhere \citep{wgan}. Previous work has demonstrated that simply constraining the spectral norm of the weight matrices with element-wise, monotonic activation functions (\textit{e.g.}, ReLU) is insufficient to maintain both expressive power and gradient norm preservation during backpropagation \citep{groupsort}.

To address this issue, researchers propose enforcing \(\mathbf{W}_l\) as orthonormal matrices for gradient norm preservation since the norm of an orthonormal matrix is exactly 1 \citep{groupsort}. There are different numerical methods to parameterize orthonormal matrices. Two widely used methods are the Björck orthonormalization algorithm \citep{bjorck,groupsort} and the Cayley transform \citep{cayley,trockman2021}. The Björck algorithm iteratively computes the closest orthonormal matrix by applying a Taylor expansion of the polar decomposition, and the Cayley transform can map a skew-symmetric matrix to an orthogonal matrix. In practice, the high-order Björck algorithm with enough iterations leads to a high-precision approximation but requires more computation, while the Cayley transform performs faster but with less precision.

\paragraph{GroupSort activation}
1-Lipschitz activation functions are common in neural networks as long as the slope is less than or equal to 1 (\textit{e.g.}, ReLU). However, to maintain the expressive power of a 1-Lipschitz neural network, the activation function should also preserve the gradient norm \citep{groupsort}. The GroupSort activation function was proposed to address this need by preserving the gradient norm. It operates by first splitting the features into groups and then sorting the features within each group to produce the output. The gradient norm-preserving property arises from the fact that its Jacobian is a permutation matrix, and permutation matrices preserve every vector norm \citep{groupsort}. Moreover, previous theoretical analyses have shown that a GroupSort neural network can represent any Lipschitz continuous piecewise linear function, making them well-suited for approximating Lipschitz continuous functions \citep{groupsort_nn}. Therefore, we employ the GroupSort function with an orthonormalized linear layer to build the 1-Lipschitz neural network.

\subsection{Model implementation}
\subsubsection{$W_1$ optimal transport solver}
As we discussed above, solving the dual problem of Wasserstein-1 optimal transport only reveals the transport direction which is not enough to recover the optimal transport map. Thus, we divide the training procedure into two steps to learn the optimal transport map.
\paragraph{Learning the Kantorovich potential}
At the first step, we aim to parameterize the 1-Lipschitz function in (\ref{eq:4}) as 1-Lipschitz neural network to maximize the dual problem (Fig. \ref{fig1}b). The training objective can be described as:

\begin{align}
    \mathcal{L}(\theta) = -\mathbb{E}_{x \sim \mu} [f_\theta(x)] + \mathbb{E}_{y \sim \nu} [f_\theta(y)]
\end{align}

where $ \{{x}_i\}_{i=1}^n $ are samples from $ \mu $, $ \{{y}_j\}_{j=1}^m $ are samples from $ \nu $, and $f_\theta: \mathbb{R}^d \mapsto \mathbb{R}$ is a 1-Lipschitz neural network parameterized by $ \theta $.

\paragraph{Learning the transportation step size}
Next, after having a fitted Kantorovich potential, we need to learn the transportation step size based on the direction of $\nabla f_\theta(x)$. However, there is not any principled method to solve this problem since the Kantorovich dual only weaken the primal problem rather than solving it in the Wasserstein-1 settings \citep{caffarelli2002constructing,trudinger2001monge}. Most recently, a method is proposed to learn a fixed step size for all samples from the data \citep{milne2022new}, but fixed step size is not accurate and will distort the desired alignment between transported and target distribution. Here, we propose to parameterize the step size function as a neural network and train it under the generative adversarial networks (GAN) paradigm to learn a sample-specific transportation. Specifically, we define a step size function $\eta_\omega: \mathbb{R}^d \mapsto \mathbb{R}^+$ parameterized by $\phi$, which takes the input sample $x$ and outputs a positive step size. If we fix the parameters of the learned potential function, then the transport map $T_{\omega}$ is then defined as (Fig. \ref{fig1}b):
\begin{align}
T_{\omega}(x) = x - \eta_\omega(x) \nabla f_\theta(x)
\end{align}
To train the step size function, we employ a discriminator network $D_\xi: \mathbb{R}^d \mapsto [0,1]$ parameterized by $\xi$, which aims to distinguish between the transported samples and the target distribution samples (Fig. \ref{fig1}b). The training objectives for the step size function and the discriminator are as follows:
\begin{align}
\mathcal{L}(\omega) &= -\mathbb{E}_{x \sim \mu}[\log D_\xi(T_{\omega}(x))] \\
\mathcal{L}(\xi) &= -\mathbb{E}_{y \sim \nu}[\log D_\xi(y)] - \mathbb{E}_{x \sim \mu}[\log(1 - D_\xi(T_{\omega}(x)))]
\end{align}
We alternate between optimizing $\mathcal{L}(\omega)$ and $\mathcal{L}(\xi)$ using stochastic gradient descent. This adversarial training process encourages the step size function to learn sample-specific transportation that aligns the transported distribution with the target distribution.

\paragraph{Comparison with $W_2$ optimal transport solver} The $W_2$ optimal transport solver typically relies on the min-max optimization of the Kantorovich dual and converges slowly \citep{w2benchmark}. Although our method avoids this step when learning the Kantorovich potential, we still employ a GAN to learn the step size function, which inherently involves a min-max optimization problem. However, we emphasize that solving the $W_1$ dual is beneficial as it simplifies the primal problem and provides direction for transportation. With this directional guidance, the GAN optimization is significantly simplified, as the generator only needs to produce a scalar step size. Thus, we believe the key advantage of our method lies in breaking down the complex optimal transport primal problem into two simpler optimization tasks.

\subsubsection{Baselines} To benchmark our method's performance, we include four baseline methods for comparison. The first baseline is the identity transformation, where no transformation is applied to the source distribution data. The second baseline is the observed target distribution, which uses the observed target distribution data from the training set for evaluation. The third baseline is scGen \citep{scgen}, which computes the displacement between the means of the source and target distributions and applies this displacement for transportation. Following the benchmark procedure in CellOT \citep{cellot}, we implement an autoencoder for scGen. Notably, we balance the training data by equalizing the number of source and target samples for scGen to avoid bias; otherwise, we observe that the performance of scGen would be worse than the identity transformation. Next, we implement the $W_2$ optimal transport solver, following the implementation in OT-ICNN \citep{makkuva2020optimal} and CellOT \citep{cellot}. Of note, CellOT is just a new PyTorch implementation of OT-ICNN, showing OT's application on single-cell datasets. Hence, we use the name ``$W_2$ OT" in our manuscript to denote these two methods. Additionally, we include CPA \citep{CPA} and biolord \citep{biolord} for the performance reference.

\subsubsection{Hyperparameters and training details}
We employ the hyperparameters used in CellOT \citep{cellot} for both the autoencoder and the $W_2$ Kantorovich potential training. Specifically, the hidden layers for the encoder and decoder are set to [512, 512] for the single-cell RNA-seq dataset (sciplex3) \citep{sciplex} and [32, 32] for the single-cell imaging dataset (4i) \citep{4i}. The latent dimension is set to 50 for the sciplex3 and 8 for the 4i. The autoencoder is trained for 250,000 iterations using the Adam optimizer with a learning rate of $1 \times 10^{-3}$ and a weight decay of $1 \times 10^{-5}$. The checkpoint with the best validation training loss will be saved and used as the final model. The $W_2$ Kantorovich potential is parameterized as an input convex neural network with hidden layers of size [64, 64, 64, 64]. The $W_2$ objective (\ref{eq:7}) is optimized over 100,000 iterations, with 10 inner iterations per optimization step, using the Adam optimizer with a learning rate of \( 1 \times 10^{-4} \) and betas (0.5, 0.9) \citep{makkuva2020optimal}.

For a fair comparison, the $W_1$ potential function employs the same hidden layer architecture as the $W_2$ model. This potential function is parameterized as a 1-Lipschitz network using the Cayley transform for fast orthonormalization approximation. According to previous work \citep{groupsort_nn}, which shows that a larger grouping size could lead to a more expressive network, we test the effect of grouping size on synthetic data \citep{w1test} and observe that a larger grouping size does not always lead to better performance within a fixed number of training iterations (Supplementary Fig. S1). The explanation for this phenomenon is that if the grouping size equals to the hidden units, the activation would degenerate into an identity transformation. Based on this observation, we set the grouping size moderately larger than 1, choosing 4 to achieve a more expressive network and to avoid the degeneration issue. This network is trained for 10,000 iterations using the Adam optimizer with betas (0.5, 0.5), along with a cosine annealing scheduler. The learning rate starts at $1 \times 10^{-2}$ and is gradually adjusted to $1 \times 10^{-4}$. To learn the step size function, both the discriminator and the step size function use hidden layers of size [64, 64, 64, 64] and are optimized for 10,000 iterations with the Adam optimizer at learning rate of $1 \times 10^{-4}$.

For the CPA model, we use the default hyperparameters provided in its repository. In our experiments, we find that these default settings perform reasonably well on the ``sciplex3" dataset but much worse on the ``4i" dataset. As there are too many hyperparameters, we then only sweep the penalty parameters and adversary steps on "4i" dataset and choose to set \texttt{penalty\_adversary=100}, \texttt{reg\_adversary=0}, and \texttt{adversary\_steps=6} for ``4i" dataset. For the biolord model, we only tune the loss penalty hyperparameters using a random perturbation condition of the ``sciplex3" dataset (1 out of 9), while keeping all other hyperparameters fixed to their reported best values. Based on the search results, we set \texttt{unknown\_attribute\_penalty=100} and \texttt{reconstruction\_penalty=100} for all experiments presented in the paper.

\subsection{Datasets and preprocessing}

\subsubsection{Synthetic toy datasets}
We first use the synthetic data from the Wasserstein-1 solver benchmark \citep{w1test} to choose the appropriate group size by testing the group sort network on synthetic data with different dimensions and different number of funnels (Supplementary Fig. S1). Next, to verify that the transportation map is ``monotonic", we generate the ``bookshelf" and ``circles" datasets. For the ``bookshelf" dataset, we created two narrow vertical strips. The source data consists of points where the x-values are uniformly distributed between 0 and 1, and the y-values are normally distributed around zero with a standard deviation of 0.001. The target data is similarly generated, but with x-values uniformly distributed between 2 and 3. This design results in two parallel strips resembling books on a shelf. For the ``circles" dataset, we used the scikit-learn \texttt{make\_circles} function to generate concentric circles with added noise. Besides evaluating the ``monotonic" property, we evaluate the nonlinearity modeling ability with the ``swiss roll" dataset, which is generated using the \texttt{make\_swiss\_roll} function. The source data is drawn from a standard 2D Gaussian distribution. The target data consists of two dimensions extracted from the 3D swiss roll and scaled appropriately. For a more complex nonlinear example, we generate the ``moons" dataset utilizing the \texttt{make\_moons} function to create two interleaving half-circles with added noise. The source and target data are composed of points from each of the two moons, respectively.

\subsubsection{Single-cell datasets} We employ the ``sciplex3" \citep{sciplex} and ``4i" \citep{4i} datasets to evaluate our method on real single-cell perturbation datasets. The ``4i" dataset consists of 37 perturbations along with a control group. It contains 48 cell imaging features after de-duplication and is less sparse compared to conventional scRNA-seq datasets. Feature selection follows the processing procedure of CellOT, which filters out the sum of the protein intensity features and retains the mean. For the ``sciplex3" scRNA-seq dataset, we normalize and log-transform the data, retaining the top 1,000 highly variable features. The processed sciplex3 data is provided by the original CellOT publication.

\subsection{Evaluation metrics}
We evaluate the effectiveness of different methods using three key metrics: the average correlation coefficient $r^2$ of feature means, the $l_2$ feature mean distance, and the maximum mean discrepancy (MMD). The $l_2$ feature means measure the Euclidean distance between the means of observed and predicted distributions, while the $r^2$ correlation coefficient quantifies the similarity between the observed and predicted means. However, relying solely on metrics that evaluate feature means can be limiting in cases where the target distribution exhibits multimodality or other complex characteristics. To effectively evaluate the performance, we employ MMD, a distributional distance metric sensitive to higher-order moments, which helps us capture discrepancies between predicted and real distributions beyond simple means \citep{mmd}. Following the implementation in CellOT, MMD is computed using an RBF kernel and averaged over several length scales to ensure robustness across varying scales of the data \citep{mmd}. 

Additionally, for high-dimensional scRNA-seq data, we conduct feature selection by identifying the top differentially expressed genes of each perturbation group using the \texttt{rank\_gene\_group} method in Scanpy \citep{scanpy}. The reason is that simply evaluating MMD in high dimensional space would lead to biased results. The top differentially expressed genes are also provided by the the original CellOT publication.

\section{Results}

\subsection{$W_1$ OT solver learns the monotonic map on 2D datasets}
\begin{figure}[!t]
\centering
\includegraphics[width=0.5\textwidth]{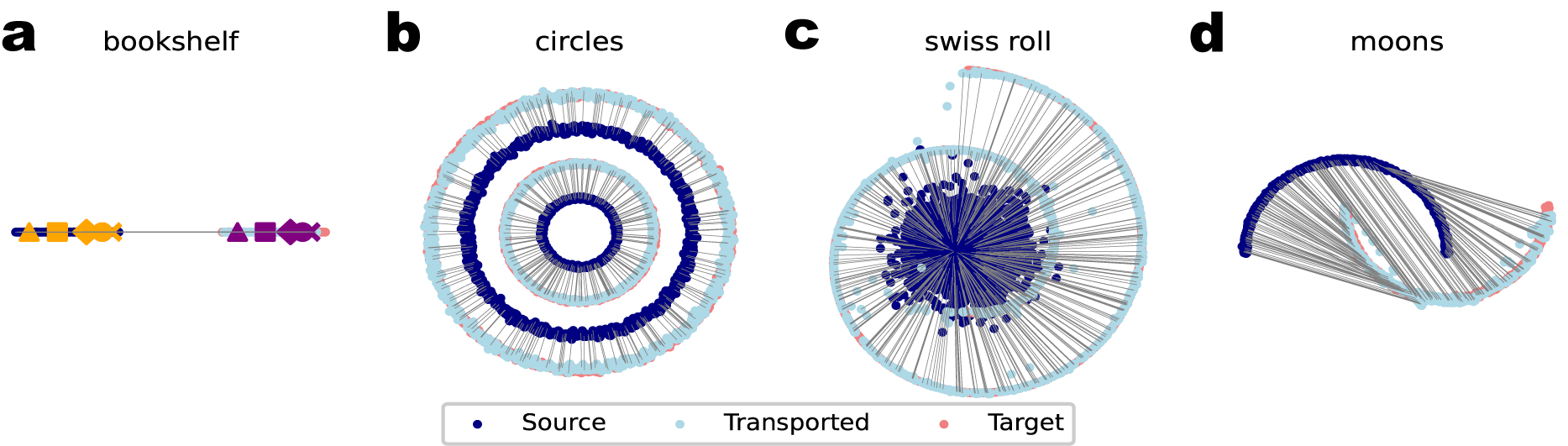}
\caption{\textbf{$W_1$OT solver on toy datasets} \textbf{a.} The ``bookshelf" datasets. To verify the transport map is ``monotonic", we set 5 markers: triangle, square, diamond, circle, and cross. The markers keep their original order after transportation. \textbf{b.} The ``ciricles" dataset. It consists of 4 concentric circles. The inner/outer source circle is expected to transported to the inner/outer target circle respectively. We can see that there is no transport between inner source circle and outer target circle. \textbf{c.} The ``swiss roll" dataset. The target distribution is a 2D swiss roll and the source distribution is a Gaussian. \textbf{d.} The ``moons" dataset. The source and target distribution are half circles.} \label{fig2}
\end{figure}
Besides the challenge of determining the transport step size in $W_1$ OT, another challenge arises from the fact that we may construct multiple transport plans solely based on the transport direction guidance. Non-uniqueness occurs in one-dimensional transport scenarios when the data points in the source and target distributions are collinear. This may cause the optimal transport solution to lose the desired ``monotonicity" property \citep{caffarelli2002constructing,feldman2002uniqueness}. In this context, ``monotonicity" refers to preserving the relative order of points after transport, as discussed in the literature. In the case of mapping two cell distributions, ``monotonicity" can be understood as preserving the local cell type structure after transport (see details in \nameref{methods}).

According to previous theoretical analyses, as long as the constructed transport map satisfies a certain condition (see Equation \ref{eq:6} in \nameref{methods}), the transport map is unique and exhibits ``monotonicity". However, this condition is intractable because it requires evaluating every pair of data points in the source distribution. Therefore, it remains a question of whether our proposed $W_1$ OT solver can find the desired ``monotonic" transport map and remain biologically meaningful in this application scenario.

To validate that our solver can learn the ``monotonic" optimal transport map, we begin by evaluating our method on two specially designed 2D datasets (see \nameref{methods}). First, we test the collinear 1D scenario, referred to as the ``bookshelf," where the source and target distributions are segments of the same length on the x-axis (Fig. \ref{fig2}a). After learning the transport direction, it is possible to construct more than one transport map with the same optimal cost. For instance, one solution could involve moving the rightmost point in the source distribution to the leftmost point in the target distribution, thus inverting the order of the entire segment. Another solution would involve shifting all points in the source distribution to the target distribution while maintaining the original order. In the experiment, we place special markers in the source distribution and observe that the relative order of the transported markers is preserved (Fig. \ref{fig2}a). Moreover, we design a 2D example consisting of concentric circles to demonstrate that our solver also learns the ``monotonic" map in a 2D scenario (Fig. \ref{fig2}b). In this example, the inner source distribution is expected to map to the inner target distribution, rather than the outer target distribution. If not, the transport map would lose its ``monotonicity" as the inner-outer relative order would be disrupted. As we can see, the learned transport map faithfully recovers the ``monotonic" map, since there are no transport rays connecting the outer source distribution to the inner target distribution (Fig. \ref{fig2}b). Next, we evaluate the $W_1$ OT solver on two difficult non-linear examples named ``swiss roll" and ``moons" (Fig. \ref{fig2}c,d). The results show that our solver can successfully learn the non-trivial transport map and exhibit similar performance to the $W_2$ OT solver in literature \citep{amos2022}.

\subsection{$W_1$ OT solver accurately predicts single-cell responses}
\begin{figure}[!t]
\centering
\includegraphics[width=0.5\textwidth]{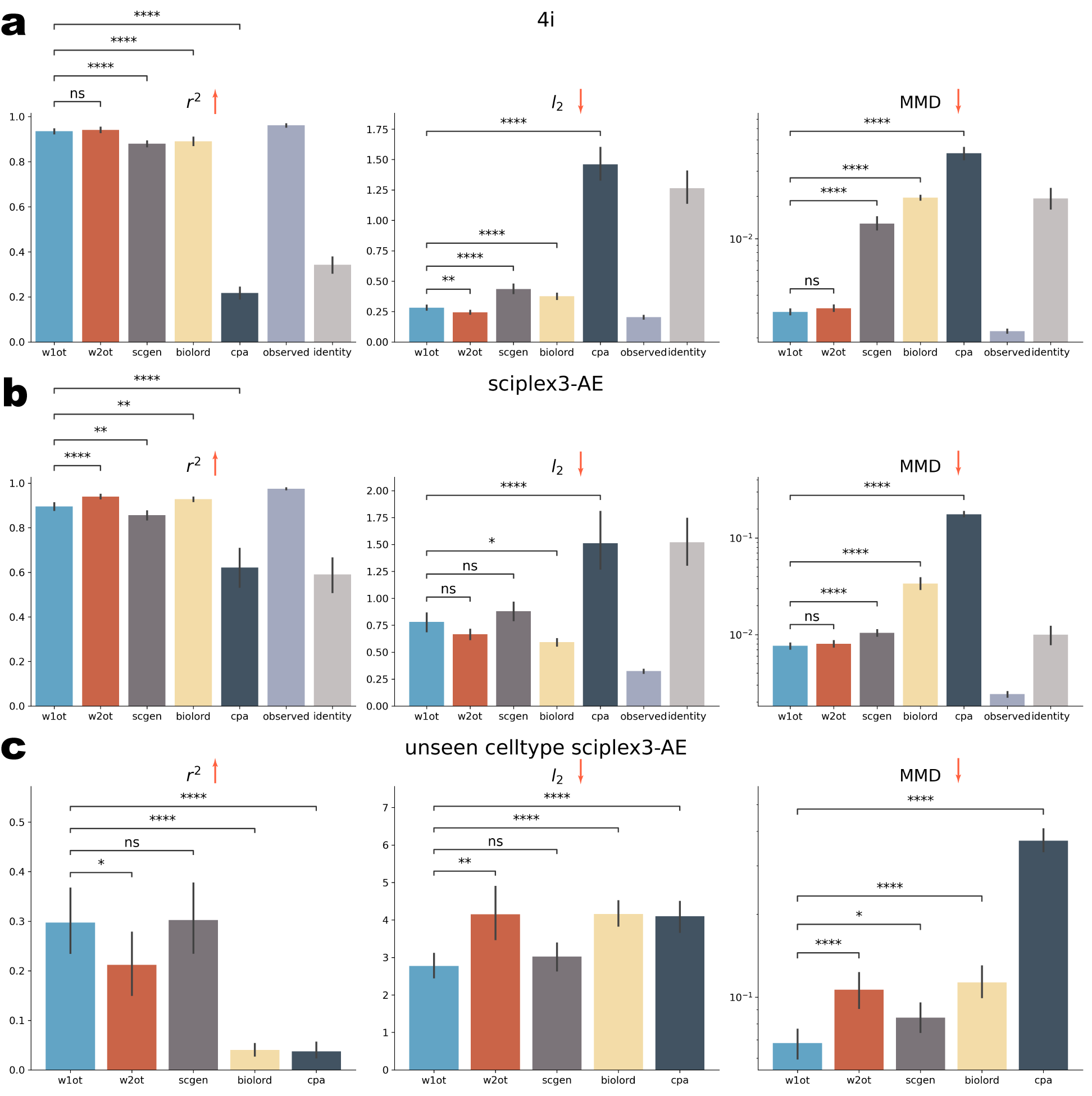}
\caption{\textbf{Performance benchmark on 2 real single-cell datasets.} \textbf{a.} The performance on 4i datasets without dimensionality reduction. Each perturbation is run independently 5 times. The performance is summarized over all perturbations in one dataset. \textbf{b.} The performance on 9 selected drugs on sciplex3 dataset in the i.i.d. setting with autoencoder embeddings and 50 latent dimensions. ``observed" denotes the best performance model can achieve. ``identity" denotes the lower bound performance. \textbf{c.} The methods' performance in the o.o.d setting, where the test set uses unseen celltype. The arrow head represents the better performance direction. All the statistical tests are Wilcoxon rank-sum test.} \label{fig3}
\end{figure}

We apply the $W_1$ OT solver to real single-cell perturbation datasets to verify its effectiveness. We also benchmark other single-cell perturbation prediction methods, including $W_2$ OT solver \citep{makkuva2020optimal,cellot}, scGen \citep{scgen}, CPA \citep{CPA}, and biolord \citep{biolord}. We are not using the chemical encoder in chemCPA \citep{chemCPA} for a fair comparison, as all other four methods do not have access to molecule information. We also report the results of directly using control group cells or perturbation group cells as prediction outputs, labeled as identity and observed, respectively, to demonstrate the lower and upper bound performance. We conduct our experiments on two datasets: the ``4i" imaging dataset \citep{cellot} and the ``sciplex3" scRNA-seq dataset \citep{sciplex}. The ``4i" imaging dataset contains 48 features for 37 perturbations while the ``sciplex3" scRNA-seq dataset contains 188 perturbations and 1,000 highly variable genes after selection. To reduce the experiments overhead, we follow the experiments in chemCPA \citep{chemCPA} to test 9 drugs: Dacinostat, Givinostat, Belinostat, Hesperadin, Quisinostat, Alvespimycin, Tanespimycin, TAK-901, and Flavopiridol, as these drugs were reported among the most effective drugs in the original publication \citep{sciplex}. For the low dimensional imaging dataset, both OT solvers are directly applied to this original feature space. While scGen is applied with latent dimension 8. For the high-dimensional scRNA-seq dataset, both OT solvers are applied to the latent space (50 features) of an autoencoder and then decoded back to cells. In the independent and identically distributed (i.i.d) setting, for each (control, perturbation) pair, we randomly split them into training and testing and repeat it 5 times for each pair. In the out-of-distribution (o.o.d) setting, since only the sciplex3 dataset provides the additional information, we split the dataset by setting a testing set composed only of unseen cell type in the training set. The unseen cell type is randomly chosen and this process is also repeated 5 times for each perturbation.

To evaluate performance, we use three metrics: $r^2$ (correlation coefficient), $l_2$ (Euclidean distance) between feature means, and Maximum Mean Discrepancy (MMD) \citep{mmd}. While $r^2$ and $l_2$ only assess differences between the means of the transported and target cell populations, MMD provides a more comprehensive evaluation by measuring the alignment of the entire distribution. For the high dimensional scRNA-seq dataset, we evaluate the top 50 differentially expressed genes between the control and perturbation groups to avoid the bias \citep{cellot}.

We report the performance of each method on both datasets in the i.i.d. setting by summarizing the results across all perturbations (Fig. \ref{fig3}a,b). We use the Wilcoxon rank-sum test to compare the performance differences between the $W_1$ OT solver and the two baselines. The results show that the $W_1$ OT solver generally achieves statistically similar performance to the $W_2$ OT solver and significantly outperforms scGen on both datasets. Furthermore, we observe that the performance of both OT solvers closely matches the observed target distribution, demonstrating their effectiveness and accuracy. Interestingly, we find that while scGen and biolord performs well on the $r^2$ and $l_2$ metrics, they are less competitive on the MMD metric. This can be attributed to the fact that scGen primarily focuses on aligning population means, but neglects the alignment of the overall distribution, as well as the fact that biolord (or CPA) is designed for latent variable disentanglement, which may not ensure proper alignment of the entire population. To further validate this assumption, we also visualize each method's embedding using UMAP \citep{umap} on the sciplex3 dataset to show the alignment of the predicted and true cell population (Supplementary Fig. S2). The results indicate that the OT method achieves better visual alignment. This observation underscores the importance of using optimal transport to map the entire distribution to the target, rather than just aligning the means.

We then test our method in the challenging o.o.d. scenario on the ``sciplex3" dataset, where both the testing source and testing target contain only an unseen cell type (Fig. \ref{fig3}c). We find that the performance of all methods drops significantly when the test data is unseen and out-of-distribution. However, our method is relatively robust compared to the baselines in terms of performance drop. Specifically, we observe that $W_1$OT outperforms $W_2$OT across all three metrics with statistical significance. This advantage of $W_1$OT could be attributed to its theoretical foundation: the higher-order $W_2$ metric is more sensitive to noise and outliers, whereas the $W_1$ metric tends to be more robust. Furthermore, we investigate the sensitivity of different methods by overexperssing some features on the ``4i" dataset.  In each run, we randomly selected 5 out of the 48 features (genes) that were expressed in at least 60\% of cells. We applied a two-fold upregulation to these genes in a subset of control cells to simulate a perturbation. The unperturbed and perturbed subsets were used as source and target data, respectively, and split into training and test sets. Models were trained to predict the perturbed gene expression profiles from the unperturbed ones. To evaluate performance, we computed the log fold change (logFC) for each perturbed gene and reported the mean absolute error between the predicted and actual logFC as the final metric (Supplementary Fig. S3). The results demonstrate that both OT-based methods are highly sensitive to gene expression changes, achieving the lowest logFC error among all models. biolord and CPA also capture the perturbation to a reasonable extent, though with higher variability. In contrast, scGen, which performs perturbation prediction in the latent space, fails to accurately recover the true fold changes, resulting in the highest error among the methods evaluated.

\subsection{$W_1$ OT solver scales better on high dimensional scRNA-seq datasets}
\begin{figure}[!t]
\centering
\includegraphics[width=0.5\textwidth]{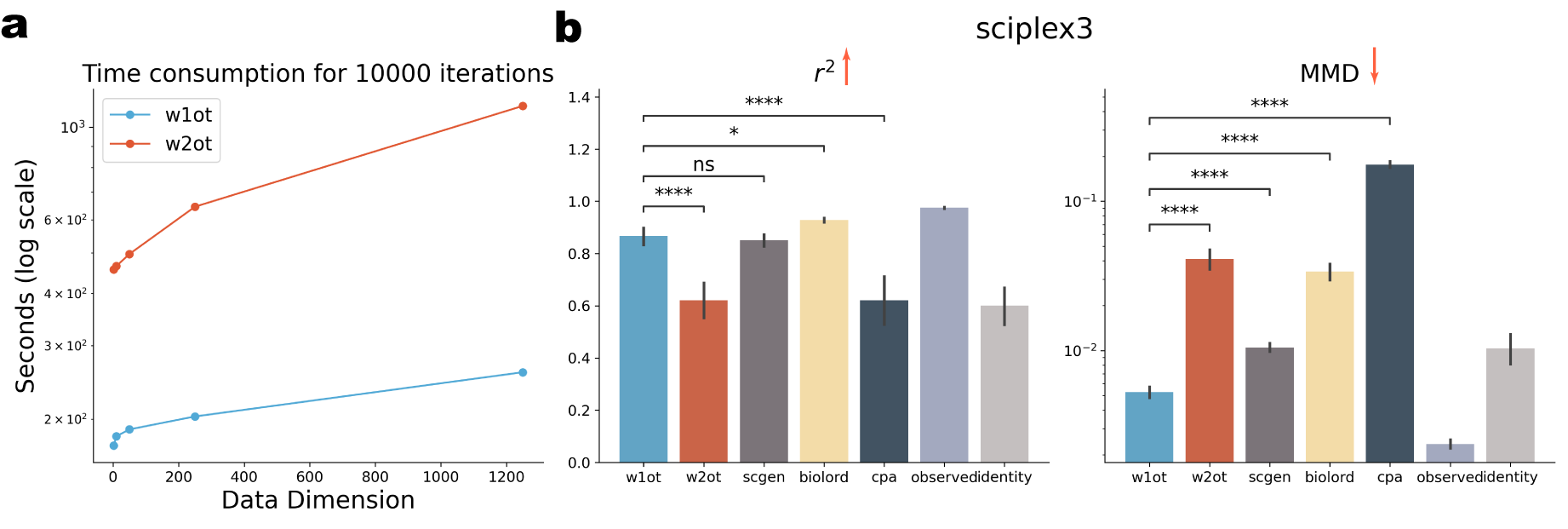}
\caption{\textbf{$W_1$OT solver on high dimensional datasets.} \textbf{a.} The scalability of two neural OT solver. It shows the time consumption of 10,000 iterations with different input data dimensionalities. \textbf{b.} The performance on sciplex3 datasets in the i.i.d. setting with 1000 highly variable genes as input for $W_1$OT and $W_2$OT, other methods use 50 latent embeddings. The arrow head represents the better performance direction. Statistical significance is shown by Wilcoxon rank-sum test.} \label{fig4}
\end{figure}
We further test the scalability of the $W_1$ OT solver by measuring its training time and performance on high-feature-dimensional datasets. To demonstrate the efficiency and speed of the $W_1$ OT solver, we compare its training time to that of the $W_2$ OT solver across various data dimensions, using an 8-core Intel Core i9-11900K processor (Fig. \ref{fig4}a). The results show that the $W_1$ OT solver is approximately 2.5 to 4.5 times faster than the $W_2$ OT solver for the same number of training iterations. Given that the default training iteration for $W_2$ OT is set at 100,000 in different studies \citep{makkuva2020optimal,cellot}, the overall speedup of $W_1$ OT would be in the range of 25 to 45 times. In practical terms, while the $W_2$ OT solver requires about an hour and a half to train, our proposed $W_1$ OT solver can complete training in just 5 minutes on a CPU. This speedup would significantly accelerate single-cell perturbation research and make training on ultra-large-scale datasets feasible.

To validate that the performance of the $W_1$ OT solver is also scalable, we applied both OT solvers directly to the ``sciplex3" dataset with 1000 features. We observe that the performance of the $W_1$ OT solver on high-dimensional data remains consistent with its performance on the latent space of 50 features (Fig. \ref{fig3} and Fig. \ref{fig4}b). However, the performance of the $W_2$ OT solver drops significantly, even falling below the lower bound (identity transformation). This indicates that solving the min-max problem is highly challenging and impractical in high-dimensional settings which has also been confirmed by previous studies \citep{w2benchmark}.

\section{Discussion}
In this work, we present a novel optimal transport solver based on the Wasserstein-1 dual formulation, which offers significant computational advantages over Wasserstein-2 neural optimal transport solvers \citep{makkuva2020optimal,w2benchmark,amos2022}. Our approach leverages 1-Lipschitz GroupSort networks combined with GANs to construct an efficient and scalable solver. Through extensive experiments, we demonstrate that our solver learns biologically meaningful ``monotonic" transport maps and achieves a $25\sim45\times$ speedup while maintaining comparable performance to existing $W_2$ OT methods on real single-cell datasets.

A key advantage of our $W_1$ OT solver is its scalability for high-dimensional data, which stems from the use of the $W_1$ dual formulation and 1-Lipschitz GroupSort networks. Although previous work questions the capability of 1-Lipschitz GroupSort networks to accurately estimate $W_1$ distance in high-dimensional settings \citep{w1test}, we demonstrate that these networks perform reasonably well on high-dimensional scRNA-seq datasets when implemented with appropriate grouping sizes \citep{groupsort_nn} (Supplementary Fig. S1). This capability is crucial as the field advances toward high-dimensional universal cell representations using single-cell foundation models \citep{geneformer,scgpt,uce}. Another computational advantage arises when we compare OT methods with diffusion or flow matching methods \citep{ddpm,flow}. Although all these methods can align two data distributions, OT infers the transported data point in one step, while diffusion or flow models infer it iteratively.

Our approach, which constructs 1-Lipschitz neural networks layer by layer, can be readily extended to model conditional transport maps $T(x, \text{condition})$. Similar to how partially input convex neural networks (PICNN) \citep{icnn} can be used to build conditional OT maps in the $W_2$ setup \citep{condot}, we can construct a partially 1-Lipschitz neural network layer $h$ as follows: $h(x,\text{condition}) = f(x) + g(\text{condition})$, where $f$ is 1-Lipschitz. With this construction, we can incorporate different conditions in the optimal transport map and enable efficient transport mapping for complex real-world applications \citep{condot}.

While we empirically validate that the $W_1$ OT solver can learn ``monotonic" maps on 2D datasets, there is currently no theoretical explanation or guarantee for this phenomenon. Additionally, although we demonstrate our solver's strong performance in various scenarios, GroupSort neural networks have not been proven to be universal 1-Lipschitz approximators under the Euclidean norm \citep{groupsort}, which may limit their performance in certain cases.

In conclusion, our $W_1$ OT solver provides a practical framework for solving the $W_1$ optimal transport problem and serves as a fast and scalable tool for single-cell perturbation prediction. Given that the $W_1$ OT solver can be integrated into other single-cell analysis tasks requiring distribution alignment, we believe it will significantly accelerate research on ultra-large-scale single-cell datasets.

\section{Competing interests}
No competing interest is declared.

\section{Author contributions statement}

Y.C. and H.H conceived the project. Y.C. built the $W_1$ OT solver framework and conducted experiments. Y.C. wrote the manuscript. Z.H. gave helpful comments on theoretical background of OT. W.C. and H.H supported the research and provided helpful comments. All authors reviewed the manuscript.

\section{Acknowledgments}
We sincerely thank the helpful discussions with Kunda Yang, Ke Ni, and Site Feng. We are very grateful for the code review work done by Ruibo Chen and Chenxi Liu. 
This work was partially supported by NSF IIS 2347592, DBI 2405416.


\bibliographystyle{abbrvnat}
\bibliography{huang.44}



\end{document}